\documentclass[conference]{IEEEtran}
\IEEEoverridecommandlockouts
\usepackage{array}
\newcolumntype{L}[1]{>{\raggedright\let\newline\\\arraybackslash\hspace{0pt}}m{#1}}
\usepackage{cite}
\usepackage{amsmath,amssymb,amsfonts}
\usepackage{algorithmic}
\usepackage{graphicx}
\usepackage{textcomp}
\usepackage{xcolor}
\usepackage{algorithm2e}
\usepackage{url}
\def\BibTeX{{\rm B\kern-.05em{\sc i\kern-.025em b}\kern-.08em
    T\kern-.1667em\lower.7ex\hbox{E}\kern-.125emX}}
\usepackage{listings}
\usepackage{multirow}
\usepackage{comment}
\usepackage{tabularx}
\usepackage{enumitem}
\setlist{leftmargin=2mm}

\begin{document}

\title{Predicting Clinical Deterioration in Hospitals
}

\author{\IEEEauthorblockN{1\textsuperscript{st} Laleh Jalali}
\IEEEauthorblockA{\textit{Hitachi America, Ltd.} \\
Santa Clara, CA, USA \\
Laleh.Jalali@hal.hitachi.com}
\and
\IEEEauthorblockN{2\textsuperscript{nd} Hsiu-Khuern Tang}
\IEEEauthorblockA{\textit{Hitachi America, Ltd.} \\
Santa Clara, CA, USA\\
Hsiu-Khuern.Tang@hal.hitachi.com}
\and
\IEEEauthorblockN{3\textsuperscript{rd} Richard H. Goldstein}
\IEEEauthorblockA{\textit{RGI Informatics, LLC.} \\
Cornwallville, NY, USA \\
rgoldstein@rgi-informatics.com}
\and
\IEEEauthorblockN{4\textsuperscript{th} Joaqu\'in \'Alvarez Rodr\'iguez}
\IEEEauthorblockA{\textit{Department of Critical Care} \\
\textit{Fuenlabrada University Hospital}\\
Fuenlabrada, Madrid, Spain \\
joaquin.alvarez@salud.madrid.org}
}

\IEEEoverridecommandlockouts
\IEEEpubid{\makebox[\columnwidth]{978-1-7281-6251-5/20/\$31.00~\copyright2020 IEEE \hfill} \hspace{\columnsep}\makebox[\columnwidth]{ }}
\maketitle

\begin{abstract}

Responding rapidly to a patient who is demonstrating signs of imminent
clinical deterioration is a basic tenet of patient care.  This gave
rise to a patient safety intervention philosophy known as a Rapid
Response System (RRS), whereby a patient who meets a pre-determined
set of criteria for imminent clinical deterioration is immediately
assessed and treated, with the goal of mitigating the deterioration
and preventing intensive care unit (ICU) transfer, cardiac arrest, or
death.  While RRSs have been widely adopted, multiple systematic
reviews have failed to find evidence of their effectiveness.
Typically, RRS criteria are simple, expert (consensus) defined rules
that identify significant physiologic abnormalities or are based on
clinical observation.

If one can find a pattern in the patient’s data earlier than the onset
of the physiologic derangement manifest in the current criteria,
intervention strategies might be more effective.  In this paper, we
apply machine learning to electronic medical records (EMR) to infer if
patients are at risk for clinical deterioration.  Our models are more
sensitive and offer greater advance prediction time compared with
existing rule-based methods that are currently utilized in hospitals.

Our results warrant further testing in the field; if successful,
hospitals can integrate our approach into their existing IT systems
and use the alerts generated by the model to prevent ICU transfer,
cardiac arrest, or death, or to reduce the ICU length of stay.

\end{abstract}


\section{Introduction}

More than~5.7 million patients are admitted annually to intensive care
units (ICUs) in the United States \cite{sccm2019}, with an estimated cost ranging from~121 to~262 billion dollars a year \cite{nates2016} and a mortality rate of about~11\% \cite{zimmerman2013}. Unplanned
admissions to the ICU are accompanied by an increase in-hospital mortality rate and length of stay.  Patients with deteriorating conditions often manifest abnormalities in physiological signals and laboratory test results before becoming unstable.  The belief that
rapid intervention in response to warning signs might provide a better
outcome for these patients gave rise to a patient safety intervention
philosophy known as a \textit{Rapid Response System} (RRS).  RRS use a
pre-determined set of criteria to identify patients at risk of
imminent clinical deterioration for immediate assessment and
treatment.  However, systematic study of these rule-based systems has
failed to show better outcomes as measured by a decrease in ICU
transfers or deaths.  There are a number of reasons that the current
RRS are inadequate: 1) they are not sensitive enough, 2) they don’t
provide sufficient lead time for medical intervention, 3) they do not
continuously monitor patients, or 4) they fail to systematically
monitor all patients \cite{alvarez2013a}.

A comprehensive, automated, timely, and accurate approach to
identifying patients who are at risk for clinical deterioration could
improve RRS intervention outcomes.  One approach is to apply machine
learning on vast historical electronic medical records (EMR) data to create state-of-the-art systems for detecting clinical deterioration; if these AI-based systems can detect patients at risk earlier and more accurately (more sensitive, no less specific), this would allow hospitals to do more timely
interventions, and potentially reduce mortality and improve prognosis.

In this study, our goal is to create an automated prediction model
based on real-time electronic medical records to identify
patients at high risk of clinical deterioration.  We use unplanned ICU
transfers as a proxy for clinical deterioration. We leverage thousands of predictor variables from EMR data, rather than dozens as is common in the current RRS. We hypothesized that such complex models provide better accuracy at longer lead time, providing more time and opportunity for clinicians to act to reverse deterioration. We compare our model
with the Spanish ICU without walls (ICUWW) method
\cite{alvarez2013}, which has been used at the same hospitals.

\begin{figure*}[htbp]
  \centering
  \includegraphics[width=5.5 in]{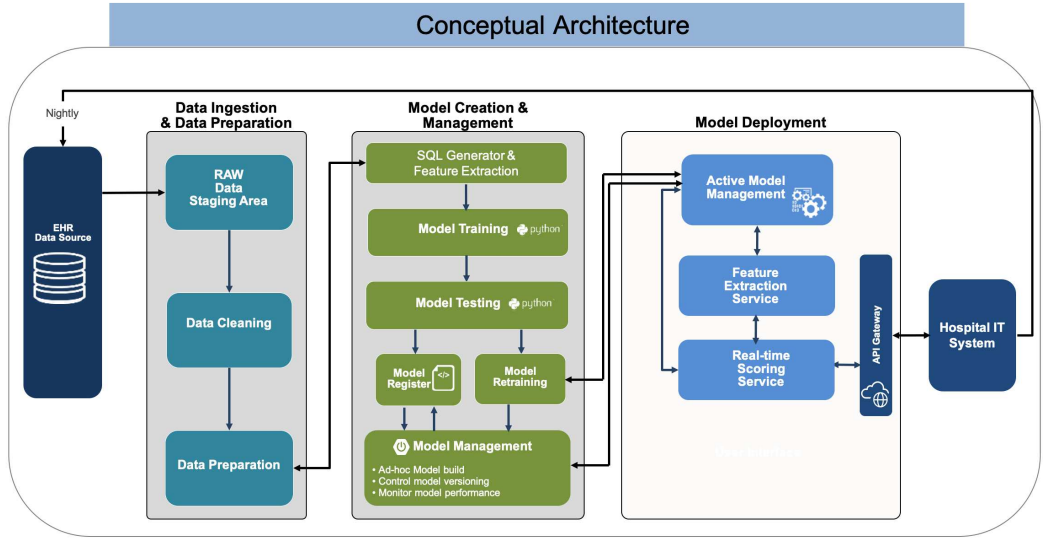}
  \caption{\strut A conceptual architecture for building and deploying machine learning solutions and big data algorithms in hospitals.}
  \label{fig:fig_conceptual_arch}
\end{figure*}

\section{Conceptual Architecture}
Figure~\ref{fig:fig_conceptual_arch} shows a conceptual architecture for building and deploying our machine learning clinical deterioration model. 

Data ingestion and preparation involves extracting data from EMR system, cleaning data, and unifying different coding standards. In practice, differences among hospitals can be substantial, making it difficult to implement and deploy an effective system across multiple hospitals. The adaptation of common coding standards by hospitals, such as LOINC for lab tests and fluids in/out, International Classification of Disease (ICD) for diagnoses and procedures, and ATC for medications, has helped to improve the usability of EMR data. Despite this, hospitals may use different standards for the same data, such as NDC and ATC for medications, or transition to a newer version of a standard, such as from ICD-9 to ICD-10. A robust system has to deal with this heterogeneity before building machine learning models. Moreover, since many EMR systems are still created and updated manually, there is room for human error and data quality issues. A proper data cleaning pipeline includes preparing data for analysis by removing or modifying data that is incorrect, incomplete, irrelevant, duplicated, or improperly formatted. 

Model creation and management involves a feature extraction and a SQL generation tool that prepares data for model building phase. Extracted features and the SQL generation tool are explained in details in section \ref{subsec:data}. 

Integrating machine learning solutions and big data algorithms to the current hospital IT system is a challenging task. A closed loop architecture indicates how predictive data shows up in clinical workflows. To integrate predictive models into different EMR systems at the point of care, we can use application program interface (API) to retrieve real-time scoring predictions . As soon as new data is available for a patient, patient information along with the new data are submitted to an API Gateway. API Gateway makes a call to the scoring service and generates a prediction score. Model deployment component is not currently addressed in our machine learning solution. It will be implemented for operationalizing the solution in future.  

\section{Research Method}
In this section, we give an overview of the data and define the
prediction target.  We formulate the real-time prediction of the
target from the streaming EMR data as a classification problem and
describe the features extracted for model learning.  We also describe
the model evaluation metrics and our modeling approach.

\subsection{Data}
\label{subsec:data}
The study population consists of about~200{,}000 admissions from three
hospitals over a period of about~5 years.  If a patient is admitted to
the hospital from the Emergency Department (ED), the ED stay is not
part of the study population. For this study, the goal is to predict unplanned ICU transfers, which after consulting with physicians at the department of critical care we use as a proxy for clinical deterioration. 

Cases in our study involved unplanned transfers from an inpatient ward to the ICU. If a patient experienced more than one ICU transfer within the same hospital admission, we only consider the first unplanned ICU transfer since the subsequent ICU transfers are considered as ICU readmission problem. The cases are identified using admission, discharge, and transfer records by selecting patients who are admitted to a general ward followed by a transfer to the ICU. We exclude ICU stays that occur at the start of an admission. This indicates a scenario where a patient is transferred to ICU from Emergency Department. Hence, we do not have access to relevant EMR data to make timely predictions for this scenario. We also exclude ICU transfers from an operating room bed or a recovery bed since those patients did not actually experience clinical deterioration, but were transferred routinely to the ICU after a surgery and those transfers are expected. Moreover, we exclude any neonatal patient data and subsequently any transfers to the neonatal intensive care unit.

Our prediction target is the first unplanned ICU transfer.  Table
\ref{tab:table1} shows the categories of patient data provided for
each admission.  Patient demographics include age and gender.  The
admission, discharge, and transfer history records the transfers of
the patient from one ward to another.  Fluids in/out, lab tests, and vital signs are collected over irregularly spaced time intervals. For all the measurements, we have access to the times at which each value was gathered. Examples of fluids in/out are
oral intake, IV intake, urine output, wound drain, and 8-hour balance, etc. that are collected two to four times a day from a patient. Some of the standard vital signs are the patient's heart rate, respiration rate, glucose level, oxygen level (SpO2), and systolic and diastolic blood pressure. Vital sign measurements are taken more frequently, ranging from 1 to 4 hours, from a patient compared to fluids in/out. In addition, there are more than 300 different lab items in the data. 

\begin{table}[ht]
  \begin{center}
    \caption{EMR data categories\strut}
    \label{tab:table1}
    \begin{tabular}{l}
      \hline \\
      Patient demographics \\ [0.15 cm]
      Admission, discharge, and transfer history\\ [0.15 cm]
      Fluids in/out\\ [0.15 cm]
      Vital signs\\ [0.15 cm]
      Lab tests\\ [0.15 cm]
      Medication orders and administration\\ [0.15 cm]
      Medications on admission\\ [0.15 cm]
      Diagnoses\\ [0.15 cm]
      Procedures\\ [0.15 cm]
      \hline
    \end{tabular}
  \end{center}
\end{table}

\subsection{Problem Formulation}

We seek to learn a model for the risk of unplanned ICU transfer from
historical patient data.  For each admission, we select a set of times
at which to calculate the risk scores, based on the arrival timestamps
of new or updated patient data, such as fluids in/out measurements,
vital signs, lab test results, and medications administered.  At each
selected time, we calculate a set of features known about the
admission at that time.

\begin{figure}[htbp]
  \centering
  \includegraphics[width=3 in]{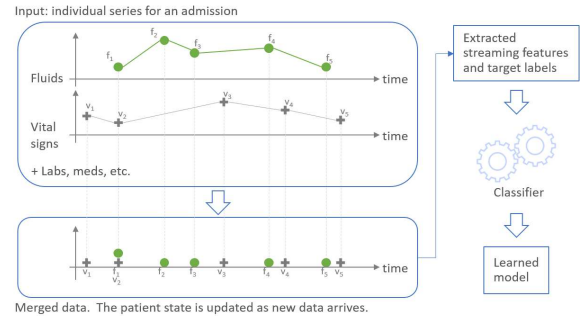}
  \caption{\strut We merge individual data series into a training
    dataset, which represents the dynamic state of a patient during
    the admission.  Features are created from the patient state and
    used to learn a model for predicting unplanned ICU transfers.}
  \label{fig:fig1}
\end{figure}

To illustrate the process, Figure~\ref{fig:fig1} shows two streams of
data for an admission.  For each fluids measurement or vital sign, the
corresponding timestamp and value is added to a patient master table.
Features for model learning are then calculated from the accumulated
data at each timestamp.  

\subsection{Feature Extraction}
Table~\ref{tab:table2} shows some of the
features that we created, including counts, averages, most recent
values, and trends, calculated over various time windows. The Charlson Comorbidity Index is a method of categorizing comorbidities of patients based on the ICD diagnosis codes \cite{deyo1992adapting}. Comorbidity means more than one disease or condition is present in the same person at the same time. Charlson comorbidity contains 17 categories, each comorbidity category has an associated weight (from 1 to 6), based on the adjusted risk of mortality or resource use, and the sum of all the weights results in a single comorbidity score for a patient. A score of zero indicates that no comorbidities were found. The higher the score, the more likely the predicted outcome will result in mortality or higher resource use. The Elixhauser Comorbidity Index is another method for measuring patient comorbidity based on ICD diagnosis codes. The original Index was developed with 30 categories \cite{elixhauser1998comorbidity}. Walraven et al. \cite{van2009modification} developed a weighting algorithm based on the association between comorbidity and death, in order to produce an overall score for the Elixhauser Index. 

The data on diagnoses, procedures, and medications on admission are only known after discharge; hence, we can only use such data from a patient’s past admissions. We calculated Charlson and Elixhauser comorbidity indexes from previous admissions of the patient, within one year. 

Some data cleaning was done to improve the quality and
interpretability of the features.  As examples: 1) we extracted the
numeric lab values from the underlying text representation, and 2) we
rounded the medication codes to the fourth level of the ATC
(Anatomical Therapeutic Chemical) hierarchy, to reduce the number of
distinct medications considered.

The combined data for all three hospitals is a matrix with about~10
million rows and~500 columns.  The rows arise from the
approximately~200{,}000 admissions and all the times within those
admissions, and the columns correspond to the features and the
prediction target, which has value~1 or~0 according to whether the
admission contains an unplanned ICU transfer or not.

\newcommand\mystrut{\rule[-0.9ex]{0pt}{3ex}}

\begin{table*}[ht]
  \begin{center}
    \caption{Examples of features for model learning\strut}
    \label{tab:table2}
\begin{tabular}{|l|l|}
\hline
Category & Features \mystrut \\ \hline
\multirow {2}{*}{Demographics} & Age \mystrut \\
                         & Gender \mystrut \\ \hline
\multirow{6}{*}{Usage recency and frequency} &  Number of previous admissions \mystrut \\
                                       &  Previous length of stay \\
                                       &  Previous discharge disposition \\ 
                                       &  Days since last discharge \\
                                       &  Number of admissions in the past $x$ months, $x \in \{1, 6, 12\}$\\ 
                                       &  Trend of the number of lab tests \mystrut \\ \hline

\multirow{2}{*}{Transfer history} &  Ward at admission \mystrut \\
                                  &  Most recent ward \mystrut \\ \hline
\multirow{5}{*}{\shortstack[l]{Fluids in/out:\\
 oral intake, IV intake,\\
 urine output, 8-hour balance, etc.}}
                                       & Most recent values \mystrut \\
                                       &  Count of the measurements in the past $y$ days, $y \in \{1, 3, 5, 7\}$\\
                                       &  Average of the measurements in the past $y$ days \\ 
                                       &  Trend of 8-hour fluid balance \\
                                       &  Trend of urine output \mystrut \\ \hline
\multirow{3}{*}{\shortstack[l]{Vital signs:\\
 heart rate, respiration rate, oxygen level (SpO2),\\
 systolic and diastolic blood pressure, glucose, etc. }}
                                       & Most recent values \mystrut \\
                                       &  Count of the measurements in the past $y$ days\\
                                       &  Average of the measurements in the past $y$ days \mystrut \\ \hline
\multirow{6}{*}{\shortstack[l]{Lab items:\\
 more than 300 lab items}}
                                       & Most recent value of each lab test \mystrut \\
                                       & Count of each lab test since admission\\
                                       & Count of each lab test in the past $y$ days\\
                                       & Average of lab test values in the past $z$ weeks, $z \in \{1, 2, 3, 4\}$ \\
                                       & Minimum of lab test values in the past $z$ weeks \\
                                       & Maximum of lab test values in the past $z$ weeks \mystrut \\ \hline
\multirow{2}{*}{Past diagnoses}   &   Modified Charlson comorbidity in the past year \mystrut \\
                                  & Modified Elixhauser comorbidity in the past year \mystrut \\ \hline
Medications                       &   Number of medications administered in the past $y$ days. \mystrut \\ \hline
\end{tabular}
  \end{center}
\end{table*}

\begin{lstlisting}[language=SQL,float=*bt, caption=Automatically generated SQL to merge features
, label= {code:sql}]
SELECT t1.facility_cd, t1.patient_id, t1.admission_id, t1.ts,
       t2.rand,
       t3.admit_ts, t3.target_ts, t3.is_direct_icu_admission,
       t4.discharge_disposition AS prev_discharge_disposition,
       t5.gender,
       t5.admission_age AS age,
       round(cast(extract(epoch FROM t1.ts - t3.admit_ts)/3600.0/24 AS numeric), 1) AS days_since_adm,
       t6.bmi AS bmi_min_1day, t6.rr AS rr_min_1day, t6.hr AS hr_min_1day, t6.spo2 AS spo2_min_1day,
       t7.bmi AS bmi_max_1day, t7.rr AS rr_max_1day, t7.hr AS hr_max_1day, t7.spo2 AS spo2_max_1day,
       ...
FROM patient_master t1
    LEFT OUTER JOIN target_adm t2 ON t1.admission_id = t2.admission_id
    LEFT OUTER JOIN adm t3 ON t1.admission_id = t3.admission_id
    LEFT OUTER JOIN adm t4 ON t3.prev_admission_id = t4.admission_id
    LEFT OUTER JOIN patient_demographics_features t5 ON t1.admission_id = t5.admission_id AND t1.ts = t5.ts
    LEFT OUTER JOIN vital_signs_min1day t6 ON t1.admission_id = t6.admission_id AND t1.ts = t6.ts
    LEFT OUTER JOIN vital_signs_max1day t7 ON t1.admission_id = t7.admission_id AND t1.ts = t7.ts
    ...
\end{lstlisting}
During deployment, the model needs to be retrained periodically to
incorporate the latest data and new hospitals that are onboarded to
the system.  For maximum flexibility, we allow the features for model
learning to reside in different tables; for example, they might be
organized by hospital and category (Table~\ref{tab:table1}).  We
developed a simple language for specifying the features to include for
modeling, their source tables, and their key columns.  We wrote a
parser for this language that generates a SQL statement for merging
those features into a single table.

The code in Listing~\ref{code:sql} shows an example of a generated SQL
statement.  Our parser supports the following attributes:
\begin{itemize}
\item Renaming (line 4)
\item Inline expressions to define new features (line 7)
\item Wildcards to select multiple features from a table (lines 8--9)
\item Custom join columns, including self-joins (lines 13--14)
\end{itemize}

Our tool makes it feasible for the end user to select features for
retraining the model without knowing the underlying details.

\subsection{Machine Learning Method}

\begin{figure}[htbp]
  \centering
  \includegraphics[width=3.2 in]{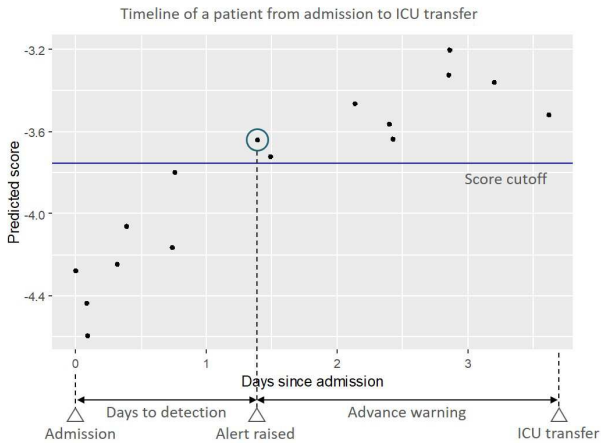}
  \caption{\strut The predicted scores at various times within an
    admission are shown.  A score is calculated when there is updated
    data for that admission.  The blue line indicates the score
    threshold.  An alert can be generated when the threshold is
    crossed for the first time.}
  \label{fig:fig2}
\end{figure}

The learned model can be used to calculate the scores for a patient
over time, based on the latest data available.  Figure~\ref{fig:fig2}
shows an example of the predicted scores for a single admission.  When
the score crosses a pre-specified threshold for the first time, an
alert can be generated to alert the medical staff to assess the
patient.  The threshold corresponds to a particular sensitivity and
specificity of this early warning system.


We evaluate model performance at the admission level.  Each admission
is either correctly or wrong classified: a false positive occurs when
an alert is generated for an admission with no unplanned ICU transfer,
and a false negative occurs when no alert is generated for an
admission with an unplanned ICU transfer.  From these, we can
calculate the sensitivity (ratio of true positives to actual
positives), specificity (ratio of true negatives to actual negatives),
and AUC of the model.  The AUC is equal to the area under the
sensitivity-specificity curve as the score threshold varies.  For true
positives, we also measure the advance warning, which is the time
between the first alert and the target event.

We selected~80\% of the admissions at random for model training and
validation.  The remaining~20\% formed the test set for model
evaluation.  We learned a separate model for each hospital, using its
own data.  We treat our problem as one of binary classification, where
for each admission and timestamp, we seek to predict whether that
admission had an unplanned ICU transfer.

We focused on tree-based models because they have natural ways of
dealing with missing data, which is prevalent in our data.  For
example, of the more than~200 possible lab tests, most are rarely
administered, leading to lots of missing values for those tests. We decided to choose XGBoost because it is modern library and readily supports a custom
evaluation metric, which we need.

We used the training and validation data to select the following model
hyperparameters: 1) the sampling rate of negative cases, 2) the
learning rate, 3) the maximum depth of the trees, 4) the sampling rate
of features used to construct each tree, and 5) the number of cases in
a node below which no further splitting is done.

For an unbalanced dataset like ours, where~99\% of the admissions do
not have an unplanned ICU transfer, sampling these negative cases is a
way to speed up model learning without sacrificing model performance.
Based on our experiments, using about~1/4 of the negative admissions
and all the positive admissions works well for model learning.  The
other four hyperparameters come from XGBoost; we selected them via a
trial-and-error approach based on the experience of XGBoost
practitioners.

\begin{figure}[htbp]
  \centering
  \includegraphics[width=3.3 in]{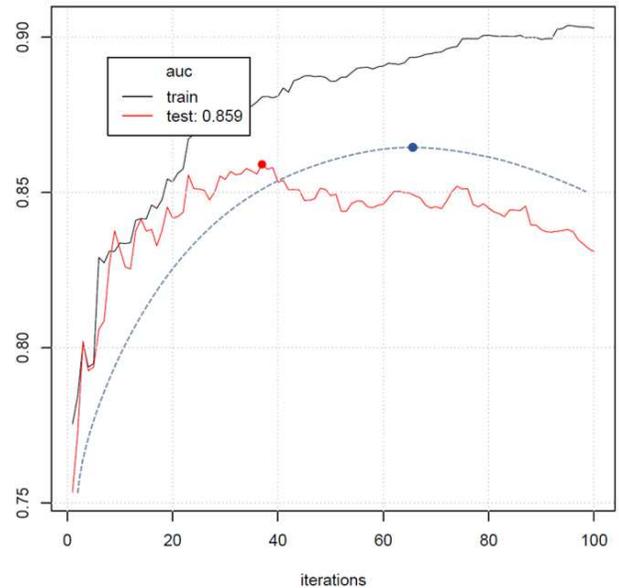}
  \caption{\strut An example showing how the number of boosting
    iterations is selected to maximize the AUC.  The default AUC (blue
    dashed line) reaches its maximum between 60 and 70 iterations, a
    region where the desired admission-level AUC (red line) is
    suboptimal.}
  \label{fig:fig3}
\end{figure}

For a given choice of the above parameters, we selected the number of
boosting iterations by 5-fold cross-validation to maximize the
\emph{admission-level} AUC.  Since the predictions are made for each
admission and timestamp, the default AUC is calculated by comparing
the predicted score $\hat y(a,t)$ for each admission~$a$ at various
timestamps~$t$ with the actual class~$y(a)$ of that admission.
However, for our purpose, each admission is either correctly
classified or not, so it is better to use the AUC obtained by
comparing the maximum predicted score $\max_t \hat y(a,t)$ within an
admission with the actual class~$y(a)$.  Figure~\ref{fig:fig3}
illustrates how using the default AUC may lead to a suboptimal choice
for the number of boosting iterations.

\section{Results}

In this section, we compare our models with the ICU without walls
(ICUWW) method \cite{alvarez2013}, which was used at the same
hospitals.  Under this method, an alert is raised if any of the lab results shown in Table \ref{tab:ICUWW_cond} are observed. We implemented these criteria for all admissions in our data, using the same set of timestamps described earlier.

\begin{table*}[ht]
  \begin{center}
    \caption{ICU Without Walls benchmark\strut}
    \label{tab:ICUWW_cond}
    \begin{tabular}{|l | l | l | l | }
      \hline
      \textbf{Lab Item} & \textbf{Lab Text Description} & \textbf{LOINC Code} & \textbf{Criteria}\\
      \hline
      ~Troponin I & Troponin I.cardiac [Mass/volume] in Serum or Plasma & 10839-9 & $>$ 0.3~ug/L \\
      \hline
      ~Arterial pH & pH of Arterial blood & 2744-1 & $<$ 7.30\\
      \hline
      ~Arterial $\mathrm{pCO}_2$ & Carbon dioxide [Partial pressure] in Arterial blood & 2019-8 & $>$60~mmHg\\
      \hline
      ~Platelets & Platelets [\#/volume] in Blood by Automated count & 777-3 & $<$ 100{,}000/uL\\
      \hline
      ~Lactate & Lactate in Serum/Plasma & 2524-7 & $>$ 3~mmol/L\\
      \hline
    \end{tabular}
  \end{center}
\end{table*}

Table~\ref{tab:table3} shows the performance of our models and of
ICUWW.  For each method, we report its sensitivity, specificity, AUC,
the median advance warning for true positives, and the mean number of
admissions per day with an alert.  For our models, these metrics were
calculated on the~20\% test set that was not used for model learning.

The ICUWW method has high specificity but low sensitivity; its average
sensitivity over the three hospitals is only~37\%.  This is perhaps
unsurprising: the~30 ``track and trigger'' scoring systems reviewed in
\cite{smith2008} have a median sensitivity of~24\% and a median
specificity of~92\%.

In Table~\ref{tab:table3}, two settings of the score threshold are
shown for each model: one with the same specificity as ICUWW and one
with a specificity of~75\%.  At the same specificity, our models are
more sensitive, have the same or greater advance warning, and raise a
similar number of alerts.  The models also have AUCs above~0.85,
compared with~0.57--0.69 for ICUWW (these are independent of the
threshold).

\begin{figure}[htbp]
  \centering
  \includegraphics[width=3.2 in]{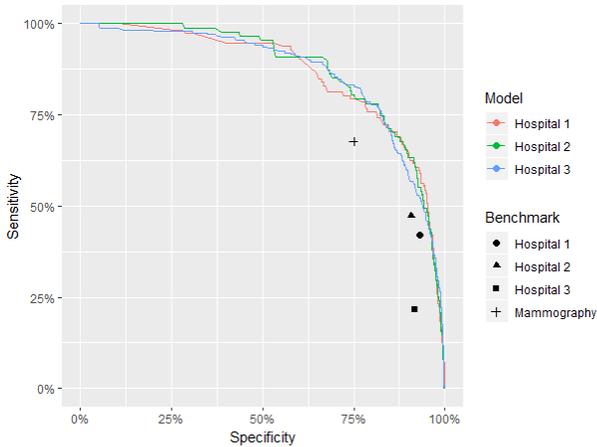}
  \caption{\strut The colored lines show the trade-off between
    sensitivity and specificity for the three hospitals.  The four
    marked points show the performance of the ICUWW benchmark and of
    mammography screening tests.}
  \label{fig:fig4}
\end{figure}

Even though the sensitivities are improved, they are still too low to
be useful.  If the model specificities are lowered to~75\%, the
sensitivities are increased substantially, to about~80-85\%, at the
cost of raising about~3 times as many alerts.  Figure~\ref{fig:fig4}
shows the sensitivities and specificities that can be attained by our
models by varying the score threshold.  Our models also compare
favorably with mammography, a routine screening test, which has a
sensitivity of~67.8\% and a specificity of~75\%.

\begin{table*}[ht]
  \begin{center}
    \caption{Performance of the models and the ICU without walls
      benchmark.  Two models are shown for each hospital, differing
      only in their score thresholds: model~A has the same specificity
      as the benchmark and model~B has a specificity of~75\%.\strut}
    \label{tab:table3}
    \begin{tabular}{l | l | r | r | r | L{1.66 cm} | L{2.19 cm}}
      \hline
      \textbf{Hospital} & \textbf{Method} & \textbf{Sensitivity} & \textbf{Specificity}
      & \textbf{AUC} & \textbf{Advance warning} &
      \textbf{Admissions per day with alerts}\\
      \hline
      1 & ICUWW & 42.0\% & 93.1\% & 0.676 & \ 2.4 days & \ \ \hphantom{0}2.9\\
      & Model A & 58.9\% & 93.1\% & 0.862 & \ 2.4 days & \ \ \hphantom{0}3.0\\
      & \strut Model B & 79.5\% & 75.0\% & 0.862 & \ 2.7 days & \ \ 11\\
      \hline
      2 & ICUWW & 47.3\% & 90.7\% & 0.690 & \ 2.1 days & \ \ \hphantom{0}3.8\\
      & Model A & 63.2\% & 90.7\% & 0.873 & \ 3.4 days & \ \ \hphantom{0}3.8\\
      & \strut Model B & 80.5\% & 75.0\% & 0.873 & \ 4.0 days & \ \ 10\\
      \hline
      3 & ICUWW & 21.6\% & 91.7\% & 0.566 & \ 0.3 days & \ \ \hphantom{0}2.4\\
      & Model A & 55.9\% & 91.7\% & 0.863 & \ 0.3 days & \ \ \hphantom{0}2.6\\
      & \strut Model B & 85.3\% & 75.0\% & 0.863 & \ 0.4 days & \ \ \hphantom{0}7.3\\
      \hline
    \end{tabular}
  \end{center}
\end{table*}

We make a final observation from Table~\ref{tab:table3}: the median
advance warning for hospital~3 is much shorter than the other two
hospitals.  We found that this is because hospital~3 sends a
significantly larger fraction of its patients to the ICU, and does so
sooner.  If we omit admissions where the ICU transfer occurred
within~24 hours of the patient being admitted, the median advance
warning for hospital~3 increases to~1.5 days for ICUWW and~2.8 days
for the model with~75\% specificity.  The average for the three
hospitals becomes~2.9 days for ICUWW and~3.6 days for the model.

\begin{figure*}[htbp]
  \centering
  \includegraphics[width=5.5 in]{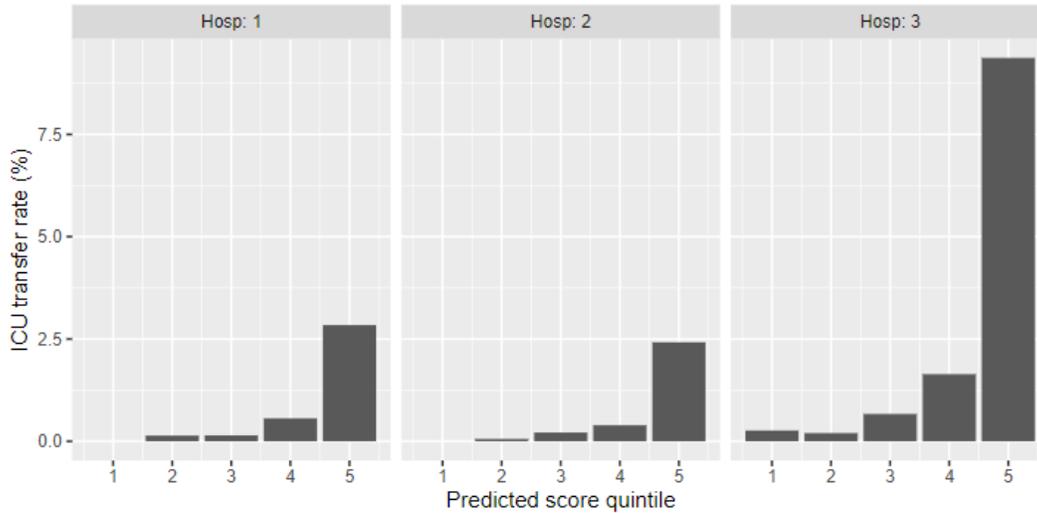}
  \caption{\strut The test set admissions for each hospital are
    divided into approximately equal-sized groups 1-5 in ascending
    order of their predicted scores.  The bars show the rates of ICU
    transfers for these groups.}
  \label{fig:fig5}
\end{figure*}

Figure~\ref{fig:fig5} shows the actual unplanned ICU transfer rates
within each predicted score quintile of the test set.  The first
quintile comprises admissions with scores in the bottom fifth of all
scores, the second quintile comprises the next fifth, and so on.  We
observe a strong correlation between the predicted score and the ICU
transfer rate.  This shows that our models are well calibrated.

\subsection{Explaining the model}

Feature importance and feature attribution are two concepts that are
useful for explaining a model.  For tree-based models, there are
appealing definitions for operationalizing these concepts, as well as
efficient calculation methods.

Feature importance measures how useful each feature is for
constructing the model or for determining the model predictions.
Suppose the model depends on the input features~$x_1$, $x_2$,
$\ldots$,~$x_p$.  For a decision tree that is constructed by
recursively splitting the training data on a single feature to
maximize the decrease in some impurity measure (such as the Gini
index), one common definition of the importance of a feature~$x_i$ is
the mean decrease in impurity over all splits on~$x_i$ weighted by the
number of training examples in those splits.  For a model that is a
sum or average of decision trees, we would simply take the sum or
average of the per-tree feature importance values.

For XGBoost, the splits are selected to minimize a different objective
function based on a loss function, and a similar definition for
feature importance can be made.

In feature attribution, we seek to allocate an individual model
prediction $\hat{y}$ to the features: $\hat{y} = \mathrm{constant} +
\phi_1 + \cdots + \phi_p$.  Recently, an attribution method based on
Shapley values has gained popularity because of its nice theoretical
properties and the availability of an efficient software
implementation for tree-based models (\cite{lundberg.erion.ea18}).
The contributions can be positive or negative, and for a
feature~$x_i$, its total absolute Shapley contribution $\sum|\phi_i|$
over the training set may be used as another definition of the
importance of~$x_i$.

We applied the above techniques to our three models, one for each
hospital.  From our analysis, \emph{age} is the most important feature
in all three models.  Some vitals signs and lab tests are also
important, although the specific ranking of these features vary across
the hospitals.  For vital signs, \emph{heart rate}, \emph{SpO2}, and
\emph{BMI} are generally important.  One surprising discovery is that
certain rarely administered lab tests are important.  They would
likely not be discovered if we had applied a sparsity criterion for
feature selection prior to model fitting. Table \ref{tab:important_features} shows the list of 30 important features, without any specific order of importance, across all three models.

\begin{table*}
\caption{The list of important features across three hospital models. The list is compiled by considering top 15 features from each model.\strut}
\label{tab:important_features}
\begin{tabularx}{\textwidth}{|>{\setlength\hsize{1.1\hsize}\setlength\linewidth{\hsize}}X|>{\setlength\hsize{0.5\hsize}\setlength\linewidth{\hsize}}X|>{\setlength\hsize{.5\hsize}\setlength\linewidth{\hsize}}X|>{\setlength\hsize{.5\hsize}\setlength\linewidth{\hsize}}X|>{\setlength\hsize{.4\hsize}\setlength\linewidth{\hsize}}X|}
\hline
\multicolumn{5}{|c|}{Classification of important features}\\
\hline
Lab tests & Vital signs & Fluids in/out & Service codes &  Demographics and Usage recency and frequency\\
\hline
\begin{itemize}
\item Lymphocytes in Blood
\item Base excess in Blood
\item Urea nitrogen in Serum or Plasma
\item Cholesterol in Serum or Plasma
\item Segmented neutrophils/100 leukocytes in Blood
\item Triglyceride in Serum or Plasma
\item Oxygen saturation in Arterial blood
\item Thyrotropin receptor Ab in Serum
\item Cancer Ag 19-9 in Serum or Plasma
\item Osmolality of Serum or Plasma
\item S100 calcium binding protein B in Serum
\item Calcium corrected for albumin in Serum or Plasma
\item S100 calcium binding protein B in Serum
\item Oxygen [Partial pressure] in Venous blood
\end{itemize} &

\begin{itemize}
\item Heart rate max 1day
\item Heart rate mrv
\item Heart rate avg
\item Systolic b.p. avg
\item Weight mrv
\item BMI mrv
\item SpO2 avg
\item Respiratory rate mrv
\end{itemize} & 

\begin{itemize}
\item IV fluid max 1day
\item 8-hour fluid balance min 1day
\end{itemize} &

\begin{itemize}
\item Digestive system is most recent service
\item Admitted to General Surgery \& Digestive System
\end{itemize} &

\begin{itemize}
\item Age
\item Previous length of stay
\item Days since prev. discharge
\item Modified Elixhauser index
\end{itemize}\\
\hline
\end{tabularx}
\end{table*}

\section{Related Work}

The Joint Commission (\url{https://www.jointcommission.org/}) is an
independent, not-for-profit organization that aims to improve health
care for the public in collaboration with other stakeholders.  A panel
of patient safety experts advises the Joint Commission on a set of
patient safety goals called the National Patient Safety Goals (NPSGs).
NPSG requires hospitals to implement systems to enable intervention
from healthcare staff when a patient's condition is worsening, which
has led to the Rapid Response System (RRS) intervention philosophy.
Some studies have shown that the implementation of RRS slightly
reduces cardiac arrest in general ward patients but does not reduce
overall hospital mortality \cite{winters2013}.

The Modified Early Warning System (MEWS) \cite{subbe2001} is a
widely used physiologic scoring system for warning about clinical
deterioration.  It is based on the patient's respiratory rate, SpO2,
temperature, systolic blood pressure, heart rate, and level of
consciousness.  MEWS typically offers an advance warning of
deterioration of~6 to~8 hours. Our benchmark ICUWW is another scoring
system; it uses five lab test results. Recent systematic reviews have demonstrated that trigger alarms only marginally improve outcomes while substantially increasing physician and nursing workloads \cite{alam14}. Moreover, these methods respond to triggering events that may not signal a truly deteriorating patient, they suffer from high rates of false alarms (70-95\%), which results in alarm fatigue and inappropriate resource utilization.

Although such rule-based systems have been widely adopted by
hospitals, multiple systematic reviews have failed to find evidence of
their effectiveness.  Typical calling criteria are based on
physiologic measurements, nurse assessments, and sometimes lab tests. Alvarez et al. \cite{alvarez2013a} trained a logistic regression model using
comprehensive EMR data to identify patients at high risk of out-of-ICU
resuscitation events and death.  They found fourteen important
predictors, including age, oxygenation, diastolic blood pressure,
arterial blood gas, lab test results, emergent orders, and assignment
to a high-risk floor.  Their automated model outperformed MEWS,
obtaining a sensitivity of~42.2\%, a specificity of~91.3\%, and an
advance warning of~15.9 hours.  Their risk scores are not calculated
in real-time; rather, data from the previous day are used to determine
the score for the current day.  In contrast, our formulation mimics
the real-time streaming of EMR data into a prediction model.

In many cases, the patients' conditions deteriorate quickly and
suffers a cardiac arrest or death outside the ICU.  Some studies
consider out-of-ICU mortality as a secondary outcome and aim to find
factors associated with it.  Wengerter et al. \cite{wengerter2018} evaluated Rothman Index variability for predicting rapid response team activation as the
primary outcome and in-hospital mortality as a secondary outcome.  The
Rothman Index incorporates twenty-six data points from EMR, including
vital signs, lab test results, cardiac rhythm, and nursing
assessments.  It is based on the contribution of each factor to a
model for predicting one-year mortality, but the Rothman Index itself
is not modeled to predict any particular outcome.  The authors
considered the Rothman Index standard deviation (RISD) and the
maximum-minus-minimum Rothman Index (MMRI) over a 24-hour window as
two measures of variability.  For predicting rapid response team
activation, RISD and MMRI have AUCs of~0.74 and~0.76, sensitivities
of~91.7\% and~92.2\%, and specificities of~39.9\% and~37.3\%,
respectively.  They found no association between these Rothman Index
variability measures and in-hospital mortality. 

Both MEWS and Rothman Index scores require manual entry by nurses and only consider small sets of data categories. Rothman Index only considers 26 data points and other work on predicting deterioration \cite{forecastICU}\cite{HASMM18} uses smaller feature sets than ours and make use of 21 predictor variables. In contrast, we have upwards of over thousands of predictor variables across six different types of EMR data categories (shown in Table \ref{tab:table1}). We believe our rich set of predictor variables not only improves the accuracy of our models but increases their robustness to missing data.

Another body of research targets ICU readmissions.  ICU readmissions
may be a more difficult target for machine learning than the first ICU
transfer since the former involves patients who have been discharged
from the ICU by a human expert (the physician) presumably after
extensive tests and monitoring. Desautels et al. \cite{desautels2017} learned a model
to predict death and 48-hour ICU readmission when a patient is first
discharged from the ICU.  Their AdaBoost model containing~1,000
decision trees was trained on the MIMIC III dataset
\cite{johnson2016} and achieved an accuracy of~70\%, a sensitivity
of~59\%, and a specificity of~66\%.

\section{Discussion}

In this paper, we built machine learning models for three hospitals to
predict unplanned transfers of patients to the ICU as a proxy for
clinical deterioration outside the ICU.  The models are tunable with a
good trade-off possibility between sensitivity and specificity.  Also,
the models take into account comprehensive data from EMR records and
achieve AUCs above~0.85.  When the specificity is set to 75\%, the
model sensitivities are around 80-85\%, with greater advance warning
compared to widely-used rule-based systems and other studies that
apply predictive models.

With its higher sensitivity and earlier warnings, our approach has the
potential to detect more patients with clinical deterioration as well
as increase the chance of a successful intervention.  The next steps
are first to determine the performance of the model in a real-world
clinical setting and then to see whether using the model in a Rapid
Response System demonstrates improved outcomes.

In addition to learning a separate model for each hospital, we also
tried learning a single model using the combined data from all three
hospitals.  The single-model results (not shown) are promising, with
only a modest decrease in performance compared with the separate
models.  Having the same model work for multiple hospitals is
important for both scalability and adoption, since it is impractical
to create a separate model for each hospital that needs such a
system---the cost of data collection and preparation alone will deter
adoption.

For a solution provider, the biggest challenge to operationalizing a
model is scalability.  The data from different hospitals are
heterogeneous and need to be normalized before they can be merged to
create a single model.  For example, the three hospitals in our study
use different names to describe their wards, which we mapped to a
common set of the most frequent names.  To scale out our approach,
significant work is needed to define common representations for all
the important EMR data elements.

In general, value co-creation is an important aspect of innovation in the healthcare industry. Before developing any AI solution, machine learning scientists and healthcare administrators should have a clear value model and continue to collaborate after the integration of the solution in the hospital's current IT system. In the case of our solution for predicting clinical deterioration, the biggest challenge for a hospital with an existing Rapid Response System is in transitioning from an intermittent system of 
patient monitoring to an always-on system using streaming data. Other issues that need to be addressed before operationalizing the solution are: 1) estimating the the efforts required to deal with hospital 
variations when we implement our system on a new hospital; 2) addressing model drift as data and clinical practice change over time; and 3) whether the value of the clinical improvements obtained by using the model outweigh the cost of implementing and maintaining the model. 

In the healthcare analytics, there is a trade-off between a more generalized prediction model that inputs big data and global features and a more specific model with a targeted use case. The features that effectively characterize a condition are the same attributes that can train an accurate predictor. But if those features do not stand out above the background noise, then the predictor only finds the noise. For this reason, predictions that are focused on a specific clinical goal will always supersede a generic predictor in terms of accuracy and utility. Hence, the full power of clinical prediction is best realized when the computational question is carefully defined, specific variables are gathered, and a targeted need is met. Our study with the carefully defined goal and attentively characterized predictors that rooted in clinical knowledge of physicians is a good example to showcase the full power of clinical prediction in an important real-life application.

\end{document}